\documentclass[10pt,journal,compsoc]{IEEEtran}
\usepackage{subcaption}
\usepackage{graphicx}
\usepackage{latexsym}
\usepackage{makecell}

\usepackage[T1]{fontenc}

\usepackage[utf8]{inputenc}

\usepackage{microtype}
\usepackage{footmisc}
\usepackage{inconsolata}
\usepackage{multirow}
\usepackage[linesnumbered,ruled,boxed]{algorithm2e} 

\usepackage{amssymb}
\usepackage{amsmath}
\usepackage{booktabs}
\usepackage{colortbl}
\usepackage{url}
\usepackage[table]{xcolor}

\ifCLASSOPTIONcompsoc
\usepackage[nocompress]{cite}
\else
\usepackage{cite}
\fi
\ifCLASSINFOpdf
\else
\fi
\begin{document}
	%
	\title{Mind Reasoning Manners: Enhancing Type Perception for Generalized 
		Zero-shot Logical Reasoning over Text}
	%
	%
	%
	%
	
	\author{Fangzhi Xu,
		Jun Liu,~\IEEEmembership{Senior Member,~IEEE,}
		Qika Lin,
		Tianzhe Zhao,
		Jian Zhang,
		and Lingling Zhang 
		\IEEEcompsocitemizethanks{\IEEEcompsocthanksitem Fangzhi Xu, Qika Lin, Tianzhe Zhao, Jian Zhang are with School
			of Computer Science and Technology, Xi'an Jiaotong University, Xi'an, China.\protect\\
			
			\IEEEcompsocthanksitem Jun Liu and Lingling Zhang are with Shaanxi Province Key Laboratory of Satellite and Terrestrial Network Tech. R\&D, National Engineering lab for Big Data Analytics, Xi'an, China.\protect\\
			
			\IEEEcompsocthanksitem The corresponding author is Jun Liu.

		}}
	
	%
	%

\markboth{Journal of \LaTeX\ Class Files,~Vol.~14, No.~8, August~2015}%
{Shell \MakeLowercase{\textit{et al.}}: Bare Demo of IEEEtran.cls for Computer Society Journals}
%



\IEEEtitleabstractindextext{%
	\begin{abstract}
		Logical reasoning task involves diverse types of complex reasoning over 
		text, based on the form of multiple-choice question answering. Given 
		the context, question and a set of options as the input, previous 
		methods achieve superior performances on the full-data setting. 
		However, the current benchmark dataset has the ideal 
		assumption that the reasoning type distribution on the train split is 
		close to the test split, which is inconsistent with many real 
		application scenarios. To address it, there remain 
		two problems to be studied: (1) How is the zero-shot capability of the 
		models (train on seen types and test on unseen types)? (2) How to 
		enhance the perception of reasoning types for the models?  For problem 
		1, we propose a new benchmark for generalized zero-shot logical 
		reasoning, named ZsLR. It includes 
		six splits based on the three type sampling strategies. For problem 2, 
		a type-aware model TaCo is proposed. It utilizes both the heuristic 
		input reconstruction and the contrastive learning to improve the type 
		perception in 
		the global representation. Extensive experiments on both the zero-shot and 
		full-data settings prove the superiority of TaCo over the state-of-the-art 
		methods. Also, we experiment and verify the generalization capability of TaCo on other logical reasoning dataset.
		
	\end{abstract}
	
	}

\maketitle

\IEEEdisplaynontitleabstractindextext

%
\IEEEpeerreviewmaketitle

\begin{figure}[t]
	\large
	\centering
	\includegraphics[scale=0.95]{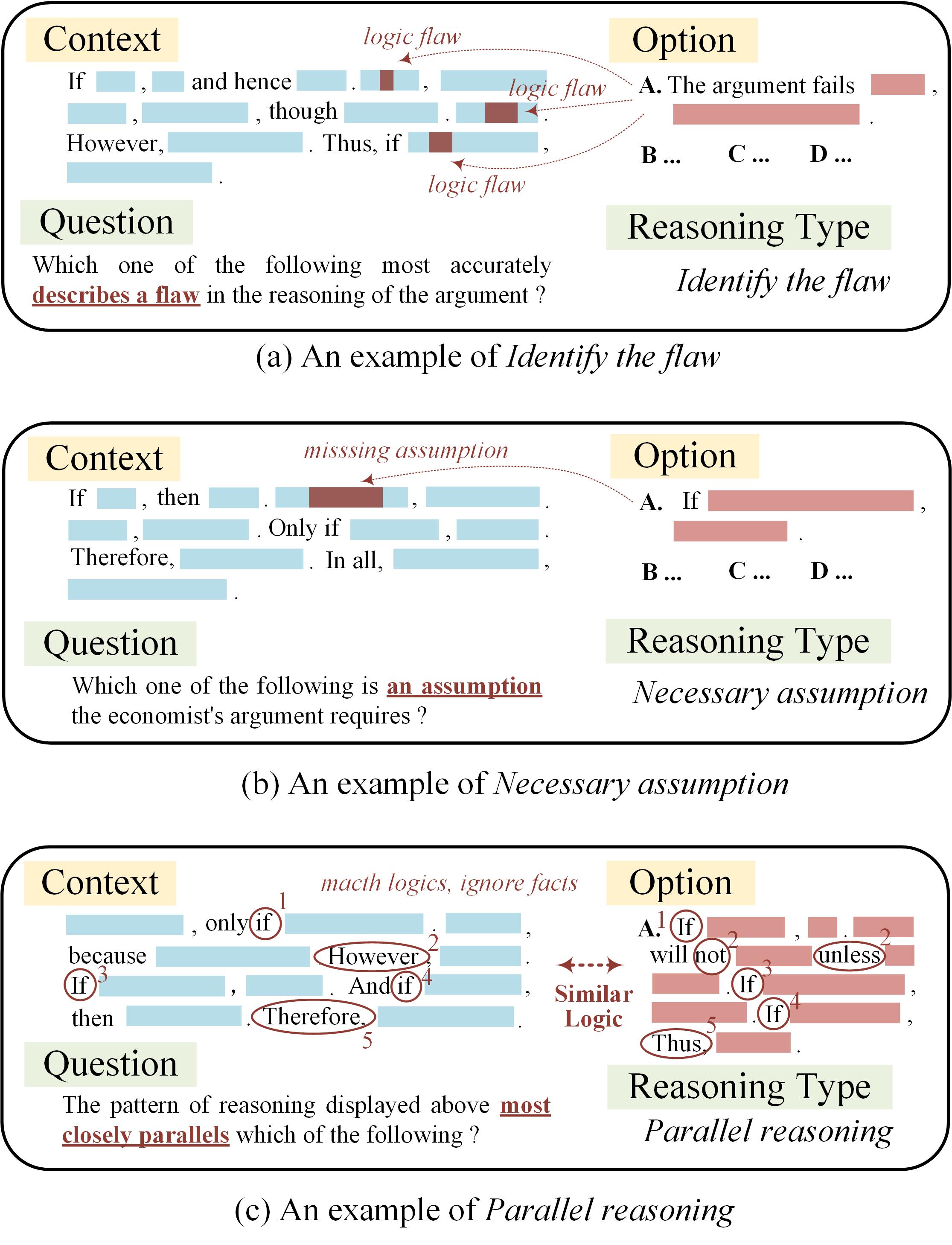}
	\caption{Examples of different reasoning types.}
	\label{fig_dataset}
	\vspace{-0.3cm}
\end{figure}

\IEEEraisesectionheading{\section{Introduction}\label{sec:introduction}}
Logical reasoning over text has aroused wide interest in the area of Machine 
Reading Comprehension (MRC) \cite{liu2019neural} and Natural Language 
Processing (NLP) \cite{hirschberg2015advances}\cite{chowdhary2020natural} recently. In the form of the 
traditional multiple-choice question answering (MCQA) \cite{richardson-etal-2013-mctest}\cite{lai-etal-2017-race}\cite{talmor-etal-2019-commonsenseqa}, the task of logical 
reasoning requires the model to perform complex reasoning and generalization. 
One of the main difficulties of the task lies in addressing diverse reasoning 
types. Fig. \ref{fig_dataset} shows some examples of reasoning types in the 
logical reasoning task. Given questions with different reasoning types, humans 
tend to focus on the respective aspects of interactions between the context and 
the option. For instance, for the type \emph{Identify the flaw} (a), the option 
is strongly related to the detailed logic flaws within the global idea. While 
for the type \emph{Necessary assumption} (b), the focus may be switched to the 
premise of the arguments and detect the missing assumption reflected by the 
option. Also, for the reasoning type of \emph{Parallel reasoning} (c), it is 
required to consider the corresponding logical structure of the context and the 
option, rather than the specific entities or events. Therefore, the modeling of 
the specific reasoning type is intuitive and necessary to the logical 
reasoning task.

Recent works have witnessed improvements in the logical reasoning tasks. 
Generally, they can be categorized into two folds: graph-based and data-based. 
In the graph-based family, DAGN \cite{huang2021dagn}, FocalReasoner 
\cite{ouyang2021fact} and Logiformer \cite{xu2022logiformer} attempt to 
construct the context graphs from different levels, such as causal and 
co-occurrence. And AdaLoGN \cite{li2022adalogn} proposes a neural-symbolic 
system in an adaptive manner. In the data-based family, previous works explore 
various data augmentation strategies. For example, LReasoner 
\cite{wang2021logic} extracts the symbols from the text and extends them with 
logical rules. MERIt \cite{jiao2022merit} designs several data generation 
methods to facilitate the training process. However, all of the methods in the 
two families lack the modeling of the reasoning type features. Although the 
above mentioned methods achieve superior results over strong baselines, the 
current researches are based on the ideal setting that train and test splits 
share the similar distribution of reasoning types. Nevertheless, in the real 
application, we are mainly exposed to the common types of questions while 
unfamiliar with the reasoning on the novel and uncommon types. In another word, 
there is an obvious gap between the ideal setting and the real scenarios. To 
address this issue, there exist two problems to be studied: 1) What about the 
test performance on the unseen types during training and how is the zero-shot 
capability of the models? 2) How to enhance the perception of reasoning types 
for the models? 

As for Problem 1, we propose a new benchmark for the generalized 
\textbf{Z}ero-\textbf{s}hot \textbf{L}ogical \textbf{R}easoning, named 
\textbf{ZsLR}. Based on the ReClor dataset \cite{yu2019reclor} with 17 
reasoning types, we form 6 zero-shot data splits according to 3 strategies 
(i.e., amount, randomness, and difficulty). To make comprehensive assessments, 
we introduce the generalized zero-shot setting \cite{pourpanah2022review}, which test on both seen types and 
unseen types with two defined metrics. The necessity and meaning of ZsLR is 
verified by the pilot experiments and it encourages more future works to mind 
reasoning manners.

As for Problem 2, we propose a \textbf{T}ype-\textbf{a}ware reasoning network 
based on \textbf{Co}ntrastive learning, named \textbf{TaCo}. First, we 
design a keyword-based extractor to output the reasoning type. Then, through 
the two designed heuristic strategies, we merge each question and the option 
into a unified \emph{Q-A} pair. Based on the co-occurrence node extraction algorithm proposed in 
Logiformer \cite{xu2022logiformer}, we form the topology within context 
part and \emph{Q-A} pair part respectively. To model the interaction of the 
context and \emph{Q-A} pair for different reasoning types, we add a global node 
to connect all the nodes of the two parts. Through the self-attention 
aggregation, the final representation of the global node is obtained and 
utilized to predict the answer. Meanwhile, we employ the sentence-BERT 
\cite{reimers2019sentence} to obtain type embeddings based on their 
descriptions. The margin loss is applied to model the reasoning type, where the 
global node serves as the anchor, ground truth type as positive example and 
other types as negative examples. In this way, the global node representation 
contains both the context and the reasoning type semantics. Therefore, the 
perception of the reasoning types can facilitate the zero-shot capability.

In all, the main contributions of this paper are summarized as follows:

(1) We are the first to focus on the issue of type-oriented reasoning 
manners. To address the issue, we propose the first benchmark: generalized 
ZsLR.\footnote{The ZsLR dataset and implementation of TaCo are public at \url{github.com/xufangzhi/TaCo}} It can well reflect the real scenarios and test the zero-shot 
capabilities of the models.

(2) We propose to tackle the zero-shot task through the heuristic 
input reconstruction and type-aware contrastive learning. The proposed model 
TaCo can function as a strong baseline for the future works on the task of ZsLR.

(3) Extensive experiments on both zero-shot and full-data settings 
prove the huge potential of the proposed issue, as well as the superiority of TaCo. Also, we conduct additional experiments to further verify the generalization capability of TaCo on other setting and dataset.

\section{Related Work}

\subsection{Logical Reasoning}
The logical reasoning task, which aims at testing the reasoning capability of 
models, has aroused wide interest. Several representative datasets have been 
proposed, such as ReClor \cite{yu2019reclor} and LogiQA \cite{liu2021logiqa}. 
Current methods on the logical reasoning task can be divided into graph-based 
methods and data-based methods. The former focuses on the construction of 
the text graphs and leverage the node connection to model the logical relations. Among this category, DAGN \cite{huang2021dagn} is the first work to split the text into EDUs and perform the reasoning with graph neural networks \cite{scarselli2008graph}\cite{welling2016semi}, but it only builds a chain-type graph and ignores the long distance interaction between nodes. FocalReasoner \cite{ouyang2021fact} proposes to attend to the fact triplets \cite{nakashole2014language} within the text and constructs a supergraph based on the extracted triplets. However, it lacks the modeling of the logical information in the text. To fully explore the logic, AdaLoGN \cite{li2022adalogn} is proposed to construct the adaptive neural-symbolic system to improve the performances. However, its reasoning process is complex and costly. Considering the above drawbacks, Logiformer \cite{xu2022logiformer} stresses much importance on both the causal relations and the co-occurrence relations in a two-branch graph transformer network. Yet it still lacks the perception of different question types, which limits the zero-shot reasoning capability of the model. The latter one is the data-based methods. These works aim to improve the performance through the data augmentation strategy. One of the representatives is LReasoner \cite{wang2021logic}, which extends the symbolic expressions by logical rules and templates. In addition, MERIt \cite{jiao2022merit} designs a meta-path-guided contrastive learning method to facilitate the training process, by utilizing the extra data. However, both of them fail to model the reasoning type of each question and lack the application value in zero-shot setting.

\subsection{Current Machine Reading Comprehension Settings}
Current machine reading comprehension (MRC) tasks \cite{zeng2020survey} can be categorized into two settings: full-data setting and low-resource (i.e., few-shot, zero-shot) setting.

Full-data setting has aroused wide concerns in the area of MRC in recent years. Popular benchmark SQuAD \cite{rajpurkar2016squad} \cite{rajpurkar2018know}, which is sourced from from Wikipedia articles, contains over 100,000 questions. It forms an abundant database for the model training. Similarly, HotpotQA \cite{yang2018hotpotqa} includes 113K questions with a variety of reasoning strategies. It stresses more on the multi-hop reasoning abilities of the model. RACE dataset \cite{lai2017race} is specially designed to improve the reading skills for both middle school and high school students, which contains about 9.7K examples. Also, some domain-specific datasets like NewsQA \cite{trischler2017newsqa}, TextbookQA \cite{kembhavi2017you}, PubMedQA \cite{jin2019pubmedqa}, aim at the area of journalism, education and biology respectively. All of them have rich training examples to improve the model capability. However, such full-data setting encourages the models to depend on the ideal scenes with abundant training data, which may not fit in some real cases.

Based on such concern, some previous works focus on the low-resource setting MRC, including few-shot or zero-shot setting. For example, to evaluate the robustness of the model, \cite{ma2021knowledge} reconstructs the data from knowledge sources for zero-shot commonsense QA. Also, \cite{bosselut2021dynamic} proposes a neuro-symbolic approach to boost the performance of zero-shot commonsense QA. Both of the methods are not exposed to the training data, forming a zero-shot setting. However, they rely on the extra knowledge bases, which limits their applications. There are also some works related to the few-shot 
QA. In \cite{ram2021few}, a new pretraining strategy is proposed to explore the realistic few-shot setting. Also, \cite{guo2021learning} tackles the few-shot 
challenge on the task of Visual Question Answering. However, all of the above methods simply obtain the few-shot splits according to the amount, which treat all the samples equally.

To sum up, current MRC benchmarks mainly focus on the common types of questions. However, in reality, it is more challenging when faced with some uncommon types of questions. To make up for these drawbacks, we attend to the reasoning types in the logical reasoning tasks and propose the first benchmark for zero-shot logical reasoning based on type attributes.

\section{The Benchmark of Generalized ZsLR}

In this section, we introduce the generalized ZsLR benchmark. At the very 
beginning, we obtain the statistical distribution of the number of 
reasoning types on the train, development and test splits, shown in Fig. 
\ref{distribution}. We arrange them in descending order of the number. The distributions are in the similar form. It 
demonstrates that the full-data setting is based on such an ideal assumption, 
which is insufficient to verify the zero-shot generalization capability of the 
models. Therefore, we propose the benchmark of generalized ZsLR and conduct the 
pilot experiments to verify the necessity.

\subsection{Zero-shot Data Construction}

\label{zero-shot_details}
The zero-shot logical reasoning datasets are split based on ReClor
\cite{yu2019reclor} without shuffling. Considering the situation in reality, 
it is easier to learn to reason on common types of samples while struggle with 
the rare ones. To this end, we design three sampling strategies, namely 
\emph{amount}, \emph{randomness} and \emph{difficulty}. 

\begin{table*}[t]
	\centering
	\caption{Details of the split datasets for the zero-shot setting. The third 
		column presents the index of the seen types and the fourth column shows the 
		number of the seen and unseen types. The last two columns present the 
		number of training samples and test samples respectively. Especially, test 
		samples contain both seen and unseen types for generalized zero-shot setting.}
	
	\begin{tabular}{|p{1.7cm}|c|cccc|}
		\hline
		\textbf{Strategy} & \textbf{Split} & \textbf{Seen Type} & 
		\textbf{\makecell[c]{\# Type \\ Seen (Unseen) }} & 
		\textbf{\makecell[c]{\# Train \\ Seen }} & 
		\textbf{\makecell[c]{\# Test \\ Seen (Unseen) }}\\
		\hline
		\hline
		\multirow{2}{*}{Amount} & v1 & \{0,3,4,8,13\} & 5 (12) & 2,190
		& 475 (525) \\
		& v2 & \{0,1,2,3,8,9,14,16\} & 8 (9) & 2,700 & 595 (405) \\
		\hline
		\multirow{2}{*}{Randomness} & v3 & \{0,2,3,13\} & 4 (13) & 1,928 & 435 
		(565)\\
		& v4 & \{0,2,3,5,7,8,13\} & 7 (10) & 2,896 & 645 (355)\\
		\hline
		\multirow{2}{*}{Difficulty} & v5 & \{0,2,4,6,8,13,15\} & 7 (10) & 2,175 
		& 473 (527)\\
		& v6 & \{1,3,5,7,9,10,11,12,14,16\} & 10 (7) & 2,463 & 527 (473)\\
		\hline
	\end{tabular}
	
	\label{tab:zeroshot_splits}
\end{table*}

For \emph{amount}, we select top-k reasoning types as the seen types, merely by 
the amount. It can be seen as a simple implementation to filter the uncommon 
types of samples.

For \emph{randomness}, we arrange the reasoning types in descending order of 
amount and select the seen ones based on the geometric distribution. The 
discrete form of the geometric distribution is,

\begin{equation}
	P\left( {X = k} \right) = {\left( {1 - p} 
		\right)^{k-1}}p,
\end{equation}
where $k$ is the sorted index of the reasoning type, $p$ is the hyper-parameter 
set to 0.1 in our implementation. In this random setting, the type with more 
training samples has a higher probability to be selected, which is in parallel 
with the real situation. 

For \emph{difficulty}, we first rank the difficulty of the reasoning types, 
based on their performance with RoBERTa-Large single model. On one split, we 
select some of the most difficult reasoning types as the seen ones. On the 
other split, we select a part of the easiest types as the seen ones.

In total, 6 zero-shot splits are obtained (2 for each strategy) for ReClor. The 
details of the splits are presented in Table \ref{tab:zeroshot_splits}.

\subsection{Generalized Zero-shot Setting}

\begin{figure}[t]
	\centering
	\begin{minipage}{\linewidth}
		\centering
		\includegraphics[scale=0.38]{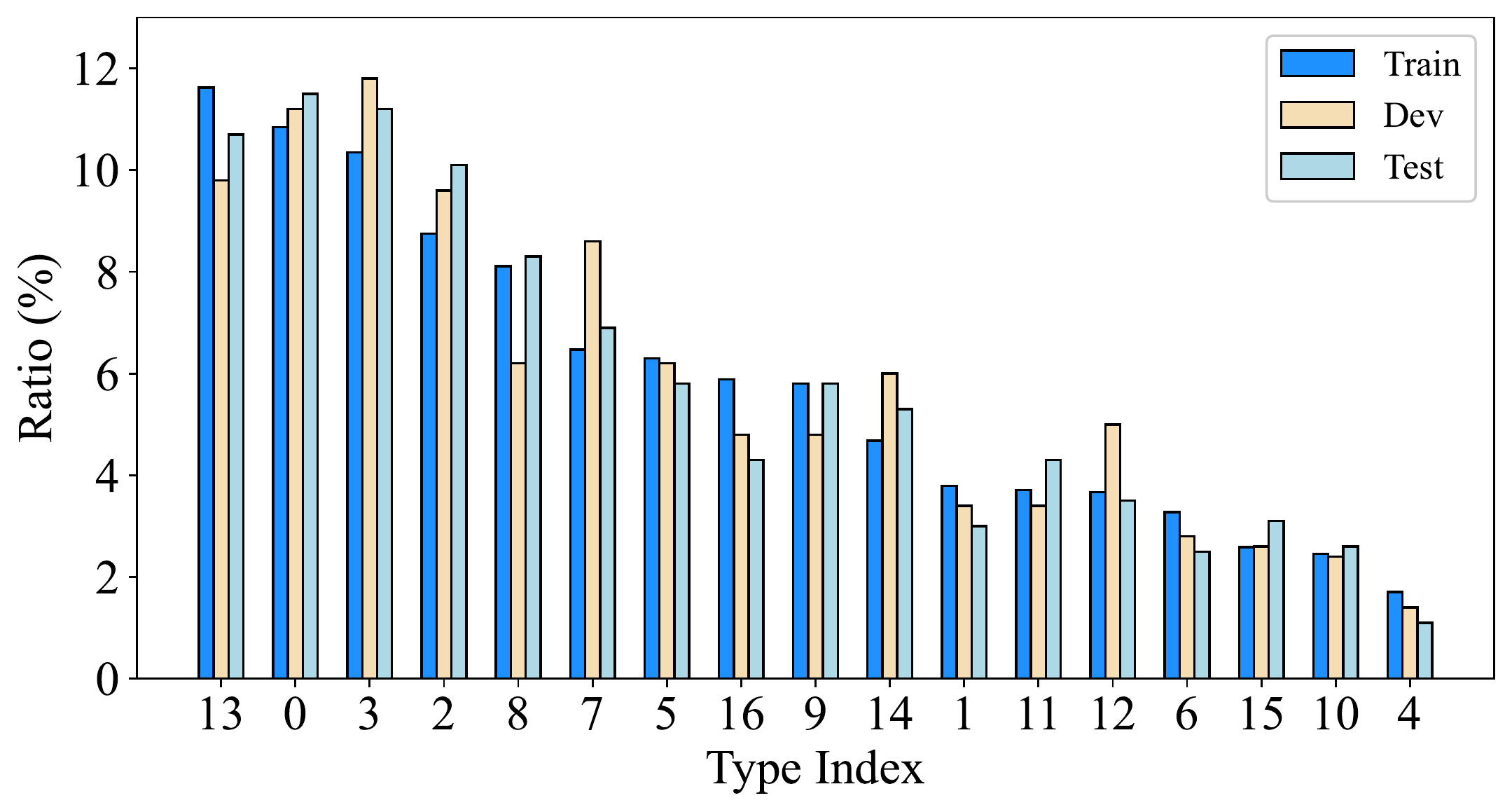}
		
		\subcaption{The distribution of the occurrence frequency of different 
			reasoning types in the ReClor dataset.}
		\label{distribution}
	\end{minipage}
	
	\begin{minipage}{\linewidth}
		\centering
		\includegraphics[scale=0.32]{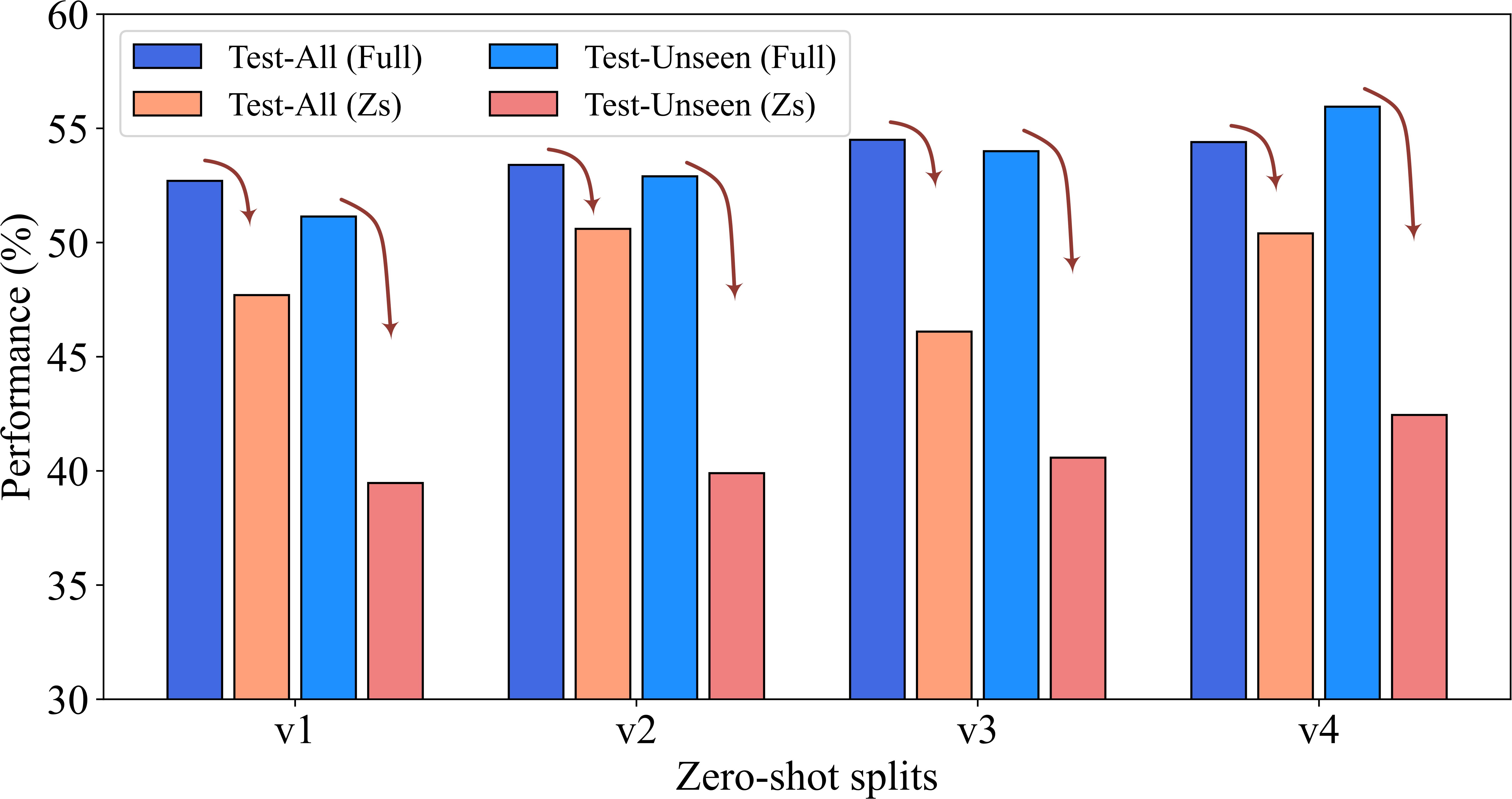}
		\subcaption{Pilot comparisons between full and zero-shot settings.}
		\label{comparison}
	\end{minipage}
	
	\caption{The visualization of the pilot experiments.}
	
	\label{pilot}
\end{figure}

We propose the zero-shot setting for the logical reasoning task. Given the  
context $C_{i}$, question $Q_{i}$ ($\emph{type}(Q_{i}) \in \mathcal{T}$) and 
option set $A_{i}$ of the $i^{th}$ question, the model is required to predict 
the correct answer $a \in A_{i}$. In the definition, $\emph{type}(Q_{i})$ 
denotes the reasoning type of the question $Q_{i}$ and $\mathcal{T}$ is the 
type set of the zero-shot splits. For each zero-shot split, a part of the types 
are sampled as the seen types, while others are viewed as the unseen ones. Only 
the questions of seen types exist during training and they consist of the 
training examples. The type set seen during the training stage and the set for the 
test stage are $\mathcal{T}^{train}$ and $\mathcal{T}^{test}$ respectively, 
thus they satisfy $\mathcal{T}^{train} \cup \mathcal{T}^{test} = \mathcal{T}$.

To be closer to the real scenes, we consider the generalized zero-shot setting. 
That is to say, the test scope is not limited to the unseen types. 
To this end, we employ two metrics for the generalized zero-shot setting, 
\emph{Test-All} and \emph{Test-Unseen}. The former one denotes the 
exact match results on the full-data test split, which contains both seen and 
unseen types, i.e., $\mathcal{T}^{train} \subseteq \mathcal{T}^{test}$. The 
latter one is the exact match results only on the unseen types of the test split, 
which ignores the performance on the seen types, i.e., $\mathcal{T}^{train} 
\cap \mathcal{T}^{test} = \emptyset$. In this way, we expect to achieve 
comprehensive assessments of the model capabilities.

\subsection{Pilot Experiments}

Also, we are going to verify the necessity of the zero-shot setting from the pilot 
experiments. For comparison with each split, we randomly sample the same amount 
of training examples on both the seen and unseen types, forming the comparison 
group. Take the zero-shot split v1 as an example, it has $2,190$ seen samples 
for training. Thus, its comparison training group also includes $2,190$ 
examples, but are distributed over all types. The purpose of the pilot 
experiments is to observe the performance changes of the two metrics on the
test split.

In the implementation, we utilize the RoBERTa single model 
\cite{liu2019roberta} to conduct the reasoning and maintain the same 
hyper-parameters of the zero-shot splits with comparison groups. Especially, to 
avoid the noise brought by the random sampling, we conduct the comparison 
experiments five times with different random seeds. Final results are the 
average values of the five experiments.

We select the results on four of the split versions for illustration (i.e., 
v1-v4), shown in Fig. \ref{comparison}. The first and the third column in 
blue are the performance of the comparison group, on the full-data test and the 
unseen types of test respectively. The second and the fourth column represent 
the performance on the zero-shot setting. With the same number of training 
examples, training only on seen types (zero-shot setting) witnesses obvious 
drops on the performance, especially on the unseen types of the test split. 

Such observations are consistent among all the zero-shot splits. Zero-shot pilot experiments based on reasoning types uncover 
the obvious drawback of the current full-data setting. In another word, it verifies the 
necessity to propose a new benchmark for the generalized ZsLR.

\section{Methods}
\begin{figure*}[t]
	\large
	\centering
	\includegraphics[scale=0.65]{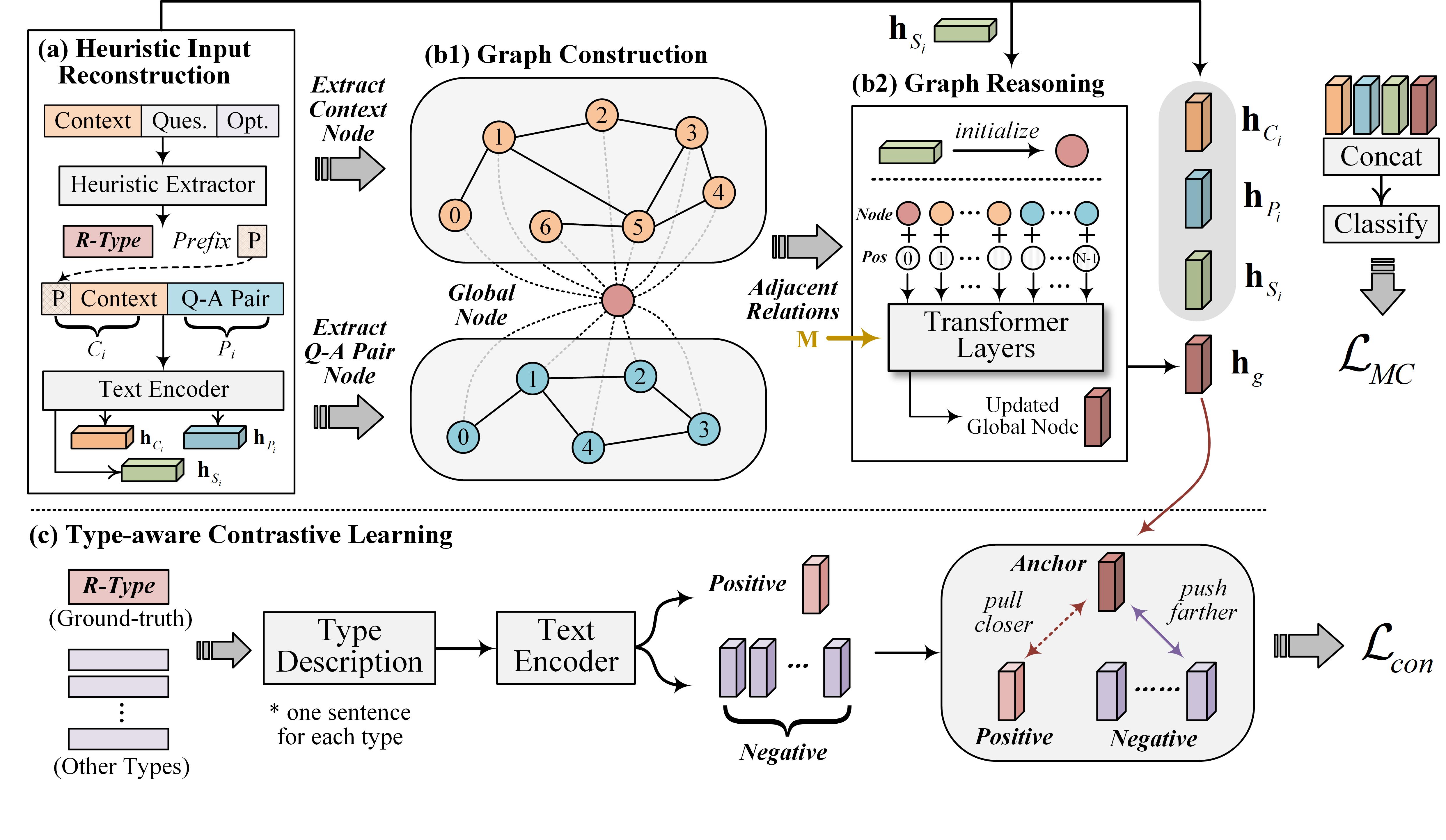}
	\vspace{-0.2cm}
	\caption{The architecture of TaCo. Taken the three basic 
		parts (i.e., context, question and option) as the inputs, we 
		reconstruct them with the heuristic strategy (a). The text graph is built 
		in module (b1) and the reasoning based on the graph transformer is 
		conducted in module (b2). The updated representation of the global node is 
		employed to perform the final inference. Meanwhile in module (c), the 
		contrastive learning is introduced to improve the awareness of the 
		reasoning type in the global node.}
	\vspace{-0.2cm}
	\label{fig_model}
\end{figure*}

In this section, we introduce the proposed methods. To tackle the zero-shot challenges, we propose a model named TaCo, which 
focuses on the reasoning type perception in the logical reasoning task. The 
architecture of TaCo is shown in Fig. \ref{fig_model}. It mainly consists of 
three parts: (a) heuristic reconstruction to acquire the type-aware input 
sequences; (b) text graph construction and reasoning for the QA problems; (c) 
type-aware contrastive learning. 

\subsection{Heuristic Input Reconstruction}
One of the common practices for MCQA problems is to concatenate three sequences 
as inputs: context, question and option. But it is insufficient for the 
zero-shot setting. There exist two main drawbacks: 1) the modeling of the 
reasoning type is implicit in the sequence; 2) it is difficult to bridge the 
interaction between context and options via the question sequence in the 
middle since it is not natural for the language model (LM). To this end, 
this paper introduces the heuristic reconstruction to the inputs based on the 
type extraction. 

To address the first drawback, we are required to label the reasoning type of 
each question based on the limited inputs. Inspired by LReasoner 
\cite{wang2021logic}, we propose a simple but effective type extractor through 
keywords. The procedure of the heuristic type extractor is presented in the pseudo-code 
of Algorithm \ref{algorithm}.

Since the ReClor dataset only makes the reasoning types on the test split 
public, we are required to collect the classification method from it. Before 
the extraction, we conclude the keywords and phrases for each reasoning type, 
forming the keywords base $\mathcal{B}$. Also, we include the maximum window 
size $\textrm{W}$ and the question sentence as the inputs. In Line 1, we first 
split the question into words and form the word sequence $S$. Then we start the 
iteration in the descending order of the window size. We slide the window over 
the question sequence to obtain the sub-sequence (Line 5). Then we match the 
sub-sequence $C$ with keywords base $\mathcal{B}_{k}$ for each reasoning type 
$k$ to derive the number of exact matches (Line 7). After each window size 
iteration, we judge the exit condition (Line 11). If there exists a unique 
type with the maximum number of matches, we label the type as the ground truth. 
Meanwhile, if we can not extract the type at the last round of iteration, we 
label the instance as \emph{Others} type (Line 16, 17).

\begin{algorithm}[t]
	\KwIn{keywords base $\mathcal{B}$, window size \textrm{W}, question 
		sentence $q$}
	\KwOut{reasoning type $T$}
	$S \leftarrow$ \texttt{SplitByWords} ($q$) \\
	\For{$i = W,...,2,1 $}
	{
		$\textrm{Count} \leftarrow \{\}$ \\
		\For{$j = 0,...,length(S)-$ \textrm{W}}
		{
			$C \leftarrow S[j:\mathrm{W}]$ \\
			\For{$k = 0,1,...,16$}
			{
				$\textrm{num} \leftarrow \texttt{NumOfMatch}(C, 
				\mathcal{B}_{k})$ \\
				Count[k] += num 
				
			}
		}
		\If{max(\textrm{Count}) > 0 and is unique}
		{
			$T \leftarrow $ type of the max match count\\
			\textbf{return}  reasoning type $T$
		}
		\If{i == 1}
		{
			\tcc{classify to `Others'}
			$T \leftarrow `Others' $ \\ 
			\textbf{return} reasoning type $T$ \\
		}
	}
	\caption{Reasoning Type Extraction}
	\label{algorithm}
\end{algorithm}

Then, we convert the type index of each question into the natural language 
(e.g., \emph{Implication}, \emph{Conclusion}). Meanwhile, to equip the LM with 
the type information, we add a type-related prefix at the beginning of the 
sequence:

\begin{equation}
	\textrm{prefix} = \textrm{This is the task of [R-Type]}
\end{equation}

\noindent where [R-Type] denotes the natural language label of the specific 
type. Thus, the type semantic is merged into the inputs in an explicit manner.

For the second drawback, one intuitive idea is to reconstruct the inputs of 
question and option sequences, transforming them into a single sequence in a 
declarative form, defined as \emph{Q-A} pair. In another word, we are 
required to fill the option into the proper position in the question. According 
to our observation, the question sentence is led by a trigger span. For 
example, in the question \emph{Which one of the following most weakens the 
	arguments above ?}, the span \emph{Which one of the following} serves as the 
trigger. Then, we can replace the trigger span with the option sequence to 
obtain the \emph{Q-A} pair, formulated as \emph{[Option] most weakens the 
	arguments above}.

Therefore, the core of the heuristic strategy is to pinpoint the precise 
position of the triggers. Although the trigger span is similar to the basic 
form of \emph{Which of the following}, it is not always the same. To further 
improve the diversity of the trigger base, we propose two heuristic strategies:

\begin{itemize}
	\item \emph{Combination}: We predefine a set of basic words (i.e., 
	\{which, one, of, the, following\}), and then randomly combine all or parts 
	of them to form the new triggers. For instances, \emph{of which the 
		following} and \emph{which of following} are two newly constructed triggers.
	
	\item \emph{Tolerance}: More trigger spans are not limited to the 
	combination of the predefined words. To this end, we propose a tolerance 
	strategy to contain the extra words within the trigger span. For example, 
	in the trigger \emph{of the following claims, which}, we include the 
	extra word \emph{claims} to the span.
\end{itemize}

In this way, the trigger base can be expanded to adapt to the complex 
situations in reality. So far, the declarative \emph{Q-A} pair is obtained.

Concatenating the context sequence $C_{i}$ and \emph{Q-A} pair sequence $P_{i}$ 
of the $i^{th}$ question, we send it into the pre-trained LM. In this paper, we 
employ the RoBERTa-Large model \cite{liu2019roberta} to obtain the token-level 
representations as follows,

\begin{equation}
	\begin{aligned}
		& {\left[\mathbf{h}_{<\mathrm{s}>} ; \mathbf{h}_{c_{0}} ; \ldots ; \mathbf{h}_{</ \mathrm{s}>} ; \mathbf{h}_{p_{0}} ; \ldots ; \mathbf{h}_{</ \mathrm{s}>}\right] } \\
		=& \operatorname{RoBERTa}\left(<\mathrm{s}>c_{0} \cdots p_{0} \cdots</ \mathrm{s}>\right),
	\end{aligned}
\end{equation}
where the tokens $\{c_{0}, c_{1}, \ldots, c_{|C_{i}|}\}$ make up $C_{i}$, and 
$\{p_{0}, p_{1}, \ldots, p_{|P_{i}|}\}$ make up $P_{i}$. The representation of 
$C_{i}$, $P_{i}$ and the whole sequence $S_{i}$ (concatenation of prefix, 
$C_{i}$ and $P_{i}$) can be obtained through the mean pooling strategy,

\begin{equation}
	\begin{aligned}
		\mathbf{h}_{C_{i}}=\frac{1}{|C_{i}|} &\sum_{k=1}^{|C_{i}|} \mathbf{h}_{c_{k}},
		\mathbf{h}_{P_{i}}=\frac{1}{|P_{i}|} \sum_{k=1}^{|P_{i}|} \mathbf{h}_{p_{k}}, \\
		&\mathbf{h}_{S_{i}}=\frac{1}{|S_{i}|} \sum_{k=1}^{|S_{i}|} 
		\mathbf{h}_{s_{k}}.
	\end{aligned}
\end{equation}

\subsection{Text Graph Construction and Reasoning}

\begin{figure}[t]
	\large
	\centering
	\includegraphics[scale=0.72]{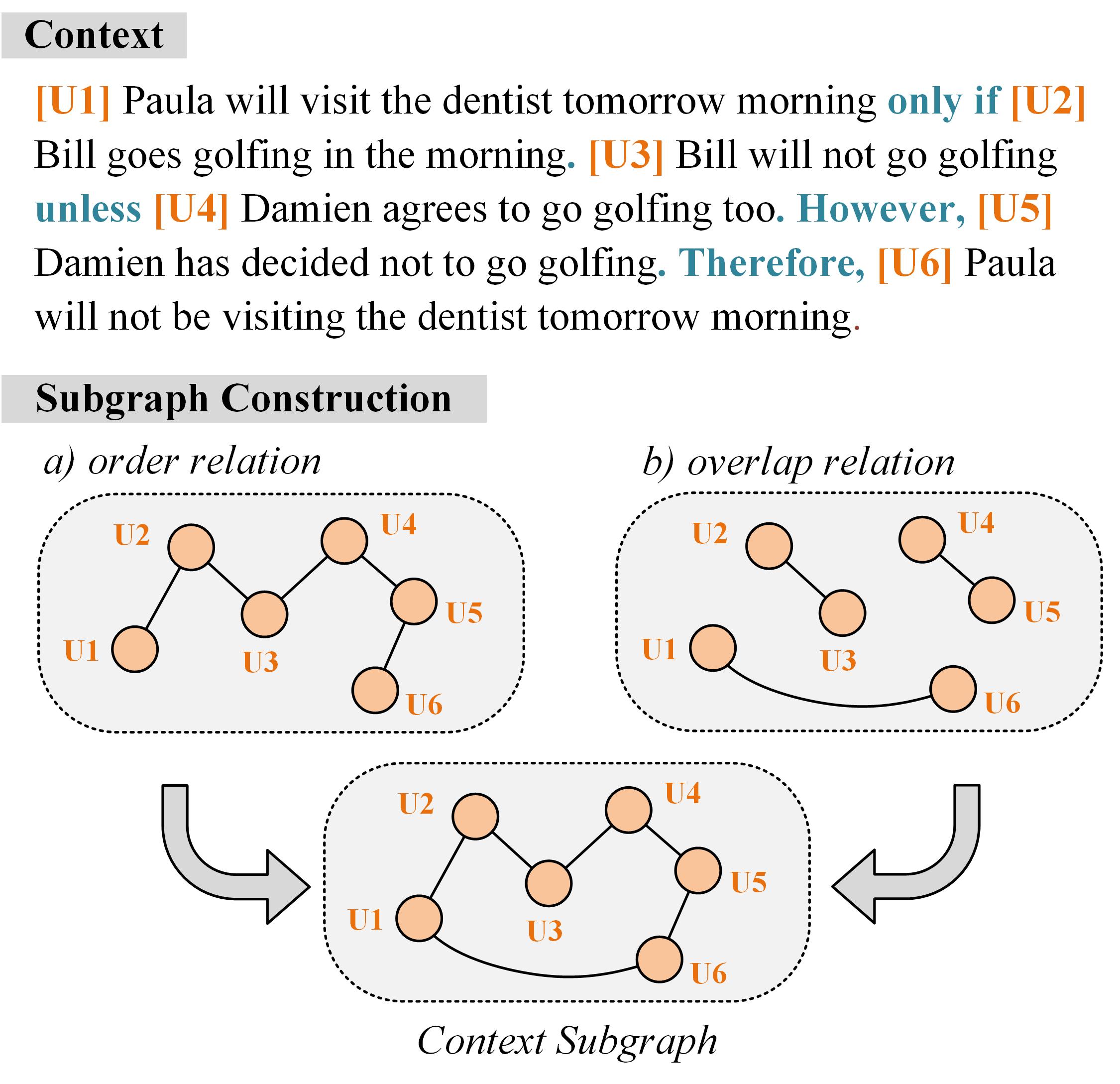}
	\caption{An example of the construction of context subgraph. The input 
		context can be split into 6 units (i.e., U1-U6). Based on the order 
		relation and overlap relation, the topology of the subgraph can be 
		obtained.}
	\label{instance}
\end{figure}

According to the previous analysis, the context and \emph{Q-A} pair may interact differently under different reasoning types. Therefore, we 
consider to build the topology structure of the two parts respectively and 
finally learn their interactive semantic. For clearer illustration, we take the 
context part as an example and present the graph construction process in 
Fig.\ref{instance}.

Firstly, we split the text sequence into units (e.g., U1-U6 in Fig. 
\ref{instance}) based on the predefined explicit connectives or punctuation 
(marked in blue in Fig. \ref{instance}). These text units function as the nodes 
during the graph reasoning process. Therefore, two node sets for the context 
and \emph{Q-A} pair parts are obtained respectively. 

To form the dual subgraphs of the input sequence, we are also required to 
define the edge relations, including a) order relation and b) overlap relation. 
The former one models the original text order, where nodes are connected in 
sequence. It forms a chain-type structure which maintains the position 
perception. The latter one models the similarity between nodes, where nodes 
with more vocabulary overlaps or similar semantics will be connected. In the 
implementation, we will transform the text units into word sets, where the 
stopwords are excluded. Through the calculation of the overlap ratio between 
sets, we connect the node pairs that exceed the threshold.

In this way, the independent topology of the context part and the \emph{Q-A} 
pair part are formed. To make the interaction of the two parts learnable, we 
add a global node to link all the extracted nodes. The global node is 
initialized with $\mathbf{h}_{S_{i}}$ and aggregates the two-way information 
flow. Therefore, this node representation is expected to contain the type 
semantic to improve the zero-shot performance.

Further, to overcome over-smoothing in graph neural networks \cite{li2018deeper} and 
fully conduct node interactions, we employ the Graph Transformer Network 
\cite{ying2021transformers, xu2022logiformer}. Let $\mathcal{N}$ 
be the node set in one graph, with the number of $|\mathcal{N}|$. We feed the 
node sequence into the Transformer architecture \cite{vaswani2017attention}. 
For each layer $l$, the updated information $\mathbf{e}_{j}$ of the node 
$\mathcal{N}_{j}$ is formulated as follows:

\begin{equation}
	\mathbf{e}_{j}^{(l)} = \sigma 
	(\mathbf{W}_{self}^{(l-1)}\mathbf{e}_{j}^{(l-1)} +\sum_{\mathcal{N}_{k} 
		\in \mathcal{N} \backslash \{\mathcal{N}_{j}\}} \alpha_{j,k}
	\mathbf{e}_{k}^{(l-1)} ). 
\end{equation} 
where $\mathcal{N} \backslash \{\mathcal{N}_{j}\}$ denotes the set without 
$j^{th}$ node.

The core of the equation is the computation for the weighted attention 
$\alpha_{j,k}$. It is obtained by the two matrices query $\mathbf{Q}$ and key 
$\mathbf{K}$:

\begin{equation}
	\mathbf{A} = \mathtt{softmax}(\frac{\mathbf{QK}^{\textrm{T}}}{\sqrt{d}} + 
	\mathbf{M}),
\end{equation}
where $\alpha_{j,k}$ is one of the elements in the matrix $\mathbf{A}$. $d$ is 
the dimension of the hidden states. $\mathbf{M} \in 
\mathbb{R}^{\mathcal{|N|}\times \mathcal{|N|}}$ is the attention bias matrix to 
encode the structural semantic of the whole graph, which is the adjacent matrix 
of the graph. 

Thus, the final feature of the global node $\mathbf{h}_{g}$ is updated through 
the graph reasoning network, containing the global semantic of the text 
sequence. Before predicting the correct answer, we concatenate 
$\mathbf{h}_{g}^{(j)}$, $\mathbf{h}_{C_{i}}^{(j)}$, $\mathbf{h}_{P_{i}}^{(j)}$ 
and $\mathbf{h}_{S_{i}}^{(j)}$ to obtain the score of the $j^{th}$ option 
followed by softmax:

\begin{equation}
	\mathrm{z}_{i}^{(j)} = \mathtt{softmax}\left(\mathbf{W} \cdot
	\left[\mathbf{h}_{g}^{(j)}; 
	\mathbf{h}_{C_{i}}^{(j)}; \mathbf{h}_{P_{i}}^{(j)}; 
	\mathbf{h}_{S_{i}}^{(j)}\right]\right),
\end{equation}
where $\mathbf{W} \in \mathbb{R}^{4d \times 1}$ denotes the linear projection. 
Same with most of the MCQA methods, we adopt the cross-entropy loss function 
for the optimization:

\begin{equation}
	\mathcal{L}_{MC} = \mathtt{CrossEntropy} \left(\mathrm{z}_{i}, 
	y_{i}\right),
\end{equation}

\noindent where $\mathrm{z}_{i}$ denotes the stack of each option score and 
$y_{i}$ is 
the ground-truth label of the $i^{th}$ example.

\begin{table*}[t]
	\centering
	\caption{The defined index and descriptions of 17 logical reasoning types 
		in the dataset.}
	\begin{tabular}{|p{0.3cm}|c|p{8.6cm}|}
		\hline
		\textbf{Id} & \textbf{Type} & \makecell[c]{\textbf{Descriptions}}\\
		\hline
		\hline
		0 & Necessary Assumptions & \makecell[l]{identify the claim that must 
			be true or is required in order for \\the argument to work.} \\
		\hline
		1 & Sufficient Assumptions & \makecell[l]{identify a sufficient 
			assumption, that is, an assumption that, if \\added to the argument, 
			would make it logically valid.}\\
		\hline
		2 & Strengthen & \makecell[l]{identify information that would 
			strengthen an argument.}\\
		\hline
		3 & Weaken & \makecell[l]{identify information that would weaken an 
			argument.}\\
		\hline
		4 & Evaluation & \makecell[l]{identify information that would be useful 
			to know to evaluate \\an argument.}\\	
		\hline
		5 & Implication & \makecell[l]{identify something that follows 
			logically from a set of premises.} \\
		\hline
		6 & Conclusion and Main Point & \makecell[l]{identify the 
			conclusion/main point of a line of reasoning.}\\
		\hline
		7 & Most Strongly Supported & \makecell[l]{find the choice that is most 
			strongly supported by a stimulus.}\\
		\hline
		8 & Explain or Resolve & \makecell[l]{identify information that would 
			explain or resolve a situation.}\\
		\hline
		9 & Principle & \makecell[l]{identify the principle, or find a 
			situation that conforms to a\\principle, or match the principles.}\\
		\hline
		10 & Dispute & \makecell[l]{identify or infer an issue in dispute.}\\
		\hline
		11 & Technique & \makecell[l]{identify the technique used in the 
			reasoning of an argument.}\\
		\hline
		12 & Role & \makecell[l]{describe the individual role that a statement 
			is playing in a \\larger argument.}\\
		\hline
		13 & Identify a Flaw & \makecell[l]{identify a flaw in an arguments 
			reasoning.}\\
		\hline
		14 & Match Flaws & \makecell[l]{find a choice containing an argument 
			that exhibits the same \\ flaws as the passages argument.}\\
		\hline
		15 & Match the Structure & \makecell[l]{match the structure of an 
			argument in a choice to the structure \\of the argument in the 
			passage.}\\
		\hline
		16 & Others & \makecell[l]{other types of questions which are not 
			included by the above.}\\
		\hline
	\end{tabular}
	
	\label{tab:type_describe}
\end{table*}

\subsection{Type-aware Contrastive Learning}
In the upper part of the architecture, the interactive semantic of the context 
and \emph{Q-A} pair is learned through one global node. Besides the implicit 
awareness of the type, we also expect the model to distinguish the ground-truth 
type from the negative ones for zero-shot logical reasoning.

Based on the heuristic extractor mentioned above, the reasoning type of each 
example can be derived. For each reasoning type, the detailed description in 
the form of natural language is utilized (listed in Table 
\ref{tab:type_describe}). 

Feeding all 17 reasoning type descriptions into the 
LM (i.e., Sentence-BERT \cite{reimers2019sentence}), we obtain the 
sentence-level embeddings, which are fixed during training. 

For each example, we propose to attend to the type information through
contrastive learning. The final representation of the global node 
$\mathbf{h}_{g}$ serves as the anchor. The ground-truth type functions as the 
positive sample, while others are negative samples, with the embedding of 
$\mathbf{h}_{gt}$ and $\mathbf{h}_{n}$ ($n$ is the index of the negative 
samples) respectively. Our purpose is to close the distance between the global 
node $\mathbf{h}_{g}$ and the positive sample $\mathbf{h}_{gt}$, while 
distancing the negative samples.

For simplicity, we model the score of two vectors using Hadamard product. 
That is:

\begin{equation}
	\begin{aligned}
		\mathrm{score}^{+} &= \mathbf{h}_{g} \odot \mathbf{h}_{gt}, \\
		\mathrm{score}^{-}_{n} &= \mathbf{h}_{g} \odot \mathbf{h}_{n}, 
	\end{aligned}
\end{equation}
where $\mathrm{score}^{+}$ and $\mathrm{score}^{-}_{n}$ represent the score of 
the positive sample and the scores of the negative ones respectively. We employ 
the margin loss as the auxiliary optimization function for each example:

\begin{equation}
	\mathcal{L}_{con} = \max \{0, \gamma - \mathrm{score}^{+} + 
	\underset{n}{\max} (\mathrm{score}^{-}_{n})\},
\end{equation}
where $\gamma > 0$ controls the difference between positive and negative 
scores. We select the maximum score of the negative one for the loss 
computation.

To maximize the joint optimization performance of the two loss functions, we 
set a trade-off coefficient $\alpha$ and transform the final loss function into:

\begin{equation}
	\mathcal{L} = \mathcal{L}_{MC} + \alpha \mathcal{L}_{con}.
\end{equation}

In this way, the global semantic for reasoning reinforces mutually with the 
type-aware semantic, which benefits the zero-shot performance. In addition, it 
provides the interpretability for the test process, which will distinguish the 
correct reasoning type from others.

\vspace{-0.3cm}
\section{Main Experiments}
In this section, we will introduce the details of the experiment setup. Also, extensive experiments on both the zero-shot and 
full-data setting will be analyzed.

\subsection{Current Full-data Benchmark}
Currently, there exist two datasets in the logical reasoning task, ReClor 
\cite{yu2019reclor} and LogiQA \cite{liu2021logiqa}. Since the LogiQA datasets 
contain only 5 reasoning types, we consider the ReClor dataset in the zero-shot
experiments. And we take both of them into account for the full-data setting to prove the generalization capability of TaCo. The detailed information of the two datasets is presented below.

\noindent \textbf{ReClor} is sourced from some standardized graduate admission 
examinations. It consists of 6,138 examples in total, including 4,638 training 
samples, 500 validation samples, and 1,000 test samples. As is listed in 
Table \ref{tab:type_describe}, ReClor contains 17 reasoning types. Especially, 
its test split is divided into two parts, which are \emph{Test-E} and 
\emph{Test-H}. The former one is the easy splits, which can be addressed 
without context. The latter one is the hard splits.

\noindent \textbf{LogiQA} is collected from National Civil Servants 
Examinations of China, including 8,678 samples in total. Among them, 7,376 are 
training samples while validation and test splits both contain 651 samples. 
Different from the diverse reasoning types in ReClor, LogiQA contains only 5 
reasoning types.

\subsection{Baselines}
We include all the previous SOTA baselines in the main paper for comparison, as 
well as the results from some classical language models. 

\noindent \textbf{Random} The results in this setting are based on the random 
predictions.

\noindent \textbf{RoBERTa-Large} \cite{liu2019roberta} The results are 
obtained simply utilizing the RoBERTa language model for the predictions. It is 
similar to the baseline of BERT-Large \cite{devlin2018bert} and XLNet-Large 
\cite{yang2019xlnet}. 

\noindent \textbf{Human Performance} \cite{yu2019reclor} In ReClor, the average 
score of different graduate students on the test split is utilized as the human 
performance.

\noindent \textbf{DAGN} \cite{huang2021dagn} It is the first work to propose 
the construction of the text graph based on the extracted EDUs. It mainly 
relies on the RoBERTa-Large \cite{liu2019roberta} to encode the tokens and 
graph neural network \cite{scarselli2008graph} to update the features.

\noindent \textbf{FocalReasoner} \cite{ouyang2021fact} It constructs a 
supergraph for reasoning, which consists of the fact units extracted from the 
text. 

\noindent \textbf{LReasoner} \cite{wang2021logic} It explores the context 
extension based on the defined logical rules (e.g., De Morgan's laws). Also, it 
employs the data augmentation method to improve the performance. Since 
constructing more training data is proved to be of great help to the zero-shot 
performance of most of the current models, we do not consider the data 
augmentation strategy when reproducing.

\noindent \textbf{MERIt} \cite{jiao2022merit} It proposes a meta-path-guided 
contrastive learning method to reason over the text. The supervised pretraining 
is performed on the abundant unlabeled text data. Due to the employment of the 
external data, we do not consider this method in the zero-shot baselines for 
the fair comparison.

\noindent \textbf{AdaLoGN} \cite{li2022adalogn} It proposes a neuro-symbolic 
system to adaptively update the text graph. Additionally, a novel 
subgraph-to-node mechanism is utilized aggregate the information.

\noindent \textbf{Logiformer} \cite{xu2022logiformer} It introduces two 
different strategies to construct the logical graph and the syntax graph 
respectively. Through the two-branch graph transformer network, the text 
features are updated to conduct the reasoning.

\begin{table}[t]
	\caption{The tuned hyper-parameters with search scopes. `GT' denotes the 
		graph transformer network.}
	\centering
	
	\begin{tabular}{|p{2.9cm}|cc|}
		\hline
		\textbf{Name of Param.} & \textbf{Search Scope} & \textbf{Best}\\
		\hline
		\hline
		\# epoch &\{10,15,20,25,30\} &30 \\
		\# head in GT &\{3,4,5,6\} &6\\
		\# layer in GT &\{3,4,5,6\} &4\\
		max sequence len &\{128,256\} &256 \\
		learning rate &\{4e-6. 5e-6, 6e-6\} &5e-6 \\
		margin $\gamma$ & \{8,10,12,14\} & 12 \\
		trade-off $\alpha$ & \{0.1,0.2,0.5,1\} & 0.2 \\
		\hline
	\end{tabular}
	
	\label{tab:tune}
\end{table}

\begin{table*}[t]
	\centering
	\caption{Experimental results on 6 zero-shot splits of ReClor 
		dataset. The percentage signs (\%) of accuracy values are omitted. The 
		optimal and sub-optimal results are marked in bold and underline 
		respectively (same for the following tables). \emph{Test-A} and 
		\emph{Test-U} denote the abbreviations of the metrics \emph{Test-All} 
		and \emph{Test-Unseen} respectively.}
	\begin{tabular}{|p{2.3cm}|cc|cc|cc|cc|cc|cc|}
		\hline
		\multirow{2}{*}{\textbf{Model}} & \multicolumn{2}{c|}{\textbf{v1}} & 
		\multicolumn{2}{c|}{\textbf{v2}} & \multicolumn{2}{c|}{\textbf{v3}} & 
		\multicolumn{2}{c|}{\textbf{v4}} & \multicolumn{2}{c|}{\textbf{v5}} & 
		\multicolumn{2}{c|}{\textbf{v6}}\\
		& Test-A & Test-U & Test-A & Test-U & Test-A & Test-U 
		& Test-A & Test-U & Test-A & Test-U 
		& Test-A & Test-U\\
		\hline
		\hline
		BERT-Large &38.00&34.36&42.00&33.39&37.50&31.61&38.00&33.26&29.60 
		&28.02&28.80&32.24 \\
		RoBERTa-Large 
		&47.70&39.47&50.60&39.90&46.10&40.58&50.40&42.45&\underline{53.00} 
		&\underline{43.66} &49.90 &50.92 \\
		DAGN&\underline{49.20}&\underline{41.37}&52.70&43.56&\underline{49.60} 
		&39.73&\underline{52.50} &44.51 &52.40 &42.63 &48.50 &49.15\\
		LReasoner  
		&46.90&40.60&50.20&43.49&48.40&\underline{42.76}&49.20&44.12&51.90 
		&42.02 &46.30&44.93\\
		Logiformer &43.50&39.31&\underline{54.80} 
		&\underline{46.30} &48.80&42.24&52.10 &\underline{44.85}
		&52.10 &40.88 &\underline{51.50} &\underline{51.44} \\
		\hline
		\hline
		TaCo 
		&\textbf{52.20}&\textbf{47.51}&\textbf{55.80}&\textbf{48.79} 
		&\textbf{52.20}&\textbf{44.26} &\textbf{54.70}&\textbf{49.89}
		&\textbf{56.00} &\textbf{46.67} &\textbf{54.70} &\textbf{55.17}\\
		\hline
	\end{tabular}
	
	\label{tab:RECLOR_zeroshot}
\end{table*}

\subsection{Implementation Details}
In this paper, all experiments run on a single GPU of Tesla A100. We employ 
RoBERTa-large model \cite{liu2019roberta} as the text encoder for all previous methods for fair comparison. For each split of the zero-shot setting, we train on the seen types and select the best epoch on the seen types of the development set for the test. We rerun other baselines on the zero-shot setting with the original configuration and maintain their reported results on the full-data setting. As for our proposed model, the hyper-parameters keep the same for both settings. The training epoch is set to 30 and the batchsize is fixed to 1. For optimization, we take Adam \cite{kingma2014adam}, with the peak learning rate as 5e-6. The number of heads and layers in the graph transformer is set to 6 and 4 respectively. $\gamma$ in the margin loss is tuned as 12. The 
loss trade-off $\alpha$ is 0.2. The selected five random seeds for the 
comparison groups are \{42, 12, 23, 234, 1234\}. Additionally, some important 
hyper-parameters are tuned for the best within a search scope. The details are 
included in the Table \ref{tab:tune}.

\subsection{Main Results}
The model TaCo is designed to bridge the gaps of the zero-shot logical 
reasoning setting, thus we are going to evaluate its performance. We present 
the results of 6 zero-shot splits in Table \ref{tab:RECLOR_zeroshot}. We rerun 
five strong baselines on the zero-shot setting, including (1) BERT-Large, (2) 
RoBERTa-Large, (3) graph-based representative method DAGN \cite{huang2021dagn}, 
(4) data-based representative model LReasoner \cite{wang2021logic}, (5) SOTA 
model Logiformer \cite{xu2022logiformer}. 

It shows that TaCo outperforms these strong baselines with large margins on all 
the zero-shot splits. Compared with the previous SOTA model Logiformer, the 
averaged improvements among all the splits are 3.80\% and 4.54\% respectively 
for the two metrics \emph{Test-All} and \emph{Test-Unseen}. And compared 
with all suboptimal results on the two metrics, TaCo still has great 
superiority of 2.50\% and 3.65\% respectively. To explore the respective roles 
of the two metrics, \emph{Test-Unseen} witnesses the greater improvements among 
all the splits, especially 6.14\% and 5.04\% over the sub-optimal methods in 
split v1 and v4 respectively. It illustrates that TaCo performs better on 
tackling the unseen types of samples than the current methods, while 
maintaining the superior performances on the seen ones. In addition, among all 
the zero-shot splits, the performances within each method vary greatly. For 
example, the previous SOTA Logiformer shows good performances in v2 but 
struggles a lot in v1. It proves that the benchmark provides diverse 
distributions of the data splits and encourages the consistent performances of 
the model. And considering the experiment results, TaCo can function as a 
strong baseline for the future work on the ZsLR benchmark.

\begin{table*}[t]
	\centering
	\caption{Ablation Studies on zero-shot splits. The drops on the 
		performances are marked in red. The darker ones represent the greater 
		decline in values.}
	\scriptsize
	\begin{tabular}{|p{4.1cm}|cc|cc|cc|cc|cc|cc|}
		\hline
		\multirow{2}{*}{\textbf{Model}} & \multicolumn{2}{c|}{\textbf{v1}} & 
		\multicolumn{2}{c|}{\textbf{v2}} & \multicolumn{2}{c|}{\textbf{v3}} & 
		\multicolumn{2}{c|}{\textbf{v4}} & \multicolumn{2}{c|}{\textbf{v5}} & 
		\multicolumn{2}{c|}{\textbf{v6}} \\
		& Test-A & Test-U & Test-A & Test-U & Test-A & Test-U 
		& Test-A & Test-U & Test-A & Test-U 
		& Test-A & Test-U\\
		\hline
		\hline
		TaCo &52.20&47.51&55.80&48.79&52.20&44.26&54.70&49.89&56.00&46.67&54.70&55.17\\
		
		\textbf{a) Input Reconstruction} & & & & & & & & & & & & \\
		\quad\quad w/o type prefix &51.10&42.30&55.30&44.01&49.90&41.08&53.80&45.60&53.60&42.04&53.50&53.67\\
		\quad\quad\quad\quad\quad$\Delta$ &\cellcolor{red!15}-1.10 
		&\cellcolor{red!45}-5.21 &\cellcolor{red!10}-0.50 
		&\cellcolor{red!45}-4.78 &\cellcolor{red!25}-2.30 
		&\cellcolor{red!35}-3.18 &\cellcolor{red!10}-0.90 
		&\cellcolor{red!45}-4.39 &\cellcolor{red!25}-2.40 &\cellcolor{red!45}-4.63 &\cellcolor{red!15}-1.20 &\cellcolor{red!15}-1.50\\
		
		\quad\quad w/o input reconstruction 
		&50.20&42.17&55.30&45.39&51.20&42.70&53.60&46.62&54.40&43.74&52.30&52.14\\
		\quad\quad\quad\quad\quad$\Delta$ &\cellcolor{red!25}-2.00 
		&\cellcolor{red!45}-5.34 &\cellcolor{red!10}-0.30 
		&\cellcolor{red!35}-3.40 &\cellcolor{red!15}-1.00 
		&\cellcolor{red!15}-1.56 &\cellcolor{red!15}-1.10 
		&\cellcolor{red!35}-3.27 &\cellcolor{red!15}-1.60 
	    &\cellcolor{red!25}-2.93 &\cellcolor{red!25}-2.40
		&\cellcolor{red!35}-3.03\\
		
		\textbf{b) Graph Construction \& Reasoning} & & & & & & & & & & & &\\
		\quad\quad w/o graph reasoning 
		&48.70&42.28&54.10&42.19&50.60&41.99&52.70&45.55&53.10&44.25 
		&50.90&50.63\\
		\quad\quad\quad\quad\quad$\Delta$ &\cellcolor{red!35}-3.50 
		&\cellcolor{red!45}-5.23 
		&\cellcolor{red!15}-1.70 &\cellcolor{red!45}-6.60 
		&\cellcolor{red!15}-1.60 
		&\cellcolor{red!25}-2.27 &\cellcolor{red!25}-2.00 
		&\cellcolor{red!45}-4.34 &\cellcolor{red!25}-2.90 
		&\cellcolor{red!25}-2.42 &\cellcolor{red!35}-3.80 
		&\cellcolor{red!45}-4.54\\
		
		\quad\quad w/o global node &51.50&46.96&53.70&44.78&51.40&43.51&53.50&46.16&55.10&46.57 
		&53.50&52.53 \\
		\quad\quad\quad\quad\quad$\Delta$ &\cellcolor{red!10}-0.70 
		&\cellcolor{red!10}-0.55 &\cellcolor{red!25}-2.10 
		&\cellcolor{red!45}-4.01 &\cellcolor{red!10}-0.80 
		&\cellcolor{red!10}-0.75 &\cellcolor{red!15}-1.20 
		&\cellcolor{red!35}-3.73 &\cellcolor{red!10}-0.90 
		&\cellcolor{red!10}-0.10 &\cellcolor{red!15}-1.20 
		&\cellcolor{red!25}-2.64\\
		
		\textbf{c) Type-aware Contrastive Learning} & & & & & & & & &&&&\\
		\quad\quad w/o type contrast 
		&51.20&42.70&55.00&44.62&50.90&43.96&54.40&46.19&55.70 &45.89&53.80&53.94 \\
		\quad\quad\quad\quad\quad$\Delta$ &\cellcolor{red!15}-1.00 
		&\cellcolor{red!45}-4.81 &\cellcolor{red!10}-0.80 
		&\cellcolor{red!45}-4.17 &\cellcolor{red!15}-1.30 
		&\cellcolor{red!10}-0.30 &\cellcolor{red!10}-0.30 
		&\cellcolor{red!35}-3.70 &\cellcolor{red!10}-0.30 
		&\cellcolor{red!10}-0.78 &\cellcolor{red!10}-0.90 
		&\cellcolor{red!15}-1.23\\
		\hline
	\end{tabular}
	\label{ablation}
\end{table*}

\subsection{Ablation Studies}

To illustrate the effectiveness of each module of TaCo in the zero-shot 
setting, we conduct the following ablation studies. The results are listed in Fig. \ref{ablation}

Detailedly, in part a), \emph{w/o type prefix} and \emph{w/o reconstruction} 
are related to the heuristic input reconstruction, where the former one ablates 
the type-oriented prefix, and the latter one maintains the simple concatenation 
of the question and option. Generally, the heuristic type prefix and input 
reconstruction bring more improvements on the metric of \emph{Test-Unseen} 
with an average of 3.95\% and 3.26\% respectively respectively on all six 
splits. It shows that type-oriented prefix is more effective than 
input reconstruction in most cases (i.e., v2-v5).

In part b), \emph{w/o graph reason} ablates the whole module of \emph{graph 
	construction and reasoning}, and replaces the global node feature with the 
pooled output of the whole sequence. Since the reasoning process is the core of 
this task, it contributes most to the performance on both the metrics, with the 
average gains of 2.58\% and 4.23\%. While \emph{w/o global node} simply 
replaces the global node feature with the pooled representation of all the 
other nodes. It can be regarded as a small part of \emph{graph construction and 
	reasoning}. From the results, it works on the splits v2, v4 and v6. As the 
training set of these three splits are obviously larger, it can be 
concluded that \emph{global node} is especially effective with more training 
samples. 

In part c), we ablate the type-aware contrastive learning module. From the 
experiments, the modeling of reasoning types does great help to the unseen 
types of samples. In the unseen types of v1, v2 and v4, the margin loss brings 
significant improvements, with gains of 4.81\%, 4.17\% and 3.70\% respectively.

Overall, the three key modules (i.e., type prefix, input reconstruction, type 
contrast) proposed to enhance the type perception all have positive effects on 
the model performances, especially on the unseen parts of the test split.

\subsection{Parameter Analysis}
\begin{figure}[t]
	\centering
	\begin{minipage}{\linewidth}
		\centering
		\includegraphics[scale=0.66]{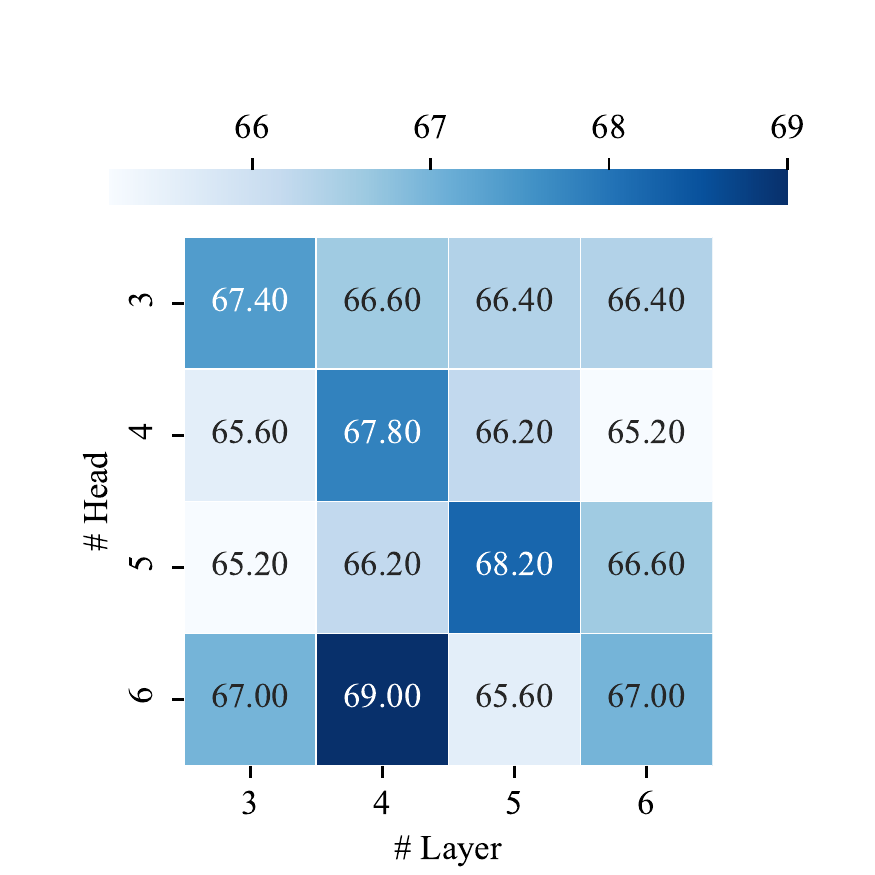}
		\subcaption{The performance on the validation split.}
		\label{heatmap_valid}
	\end{minipage}
	
	\begin{minipage}{\linewidth}
		\centering
		\includegraphics[scale=0.66]{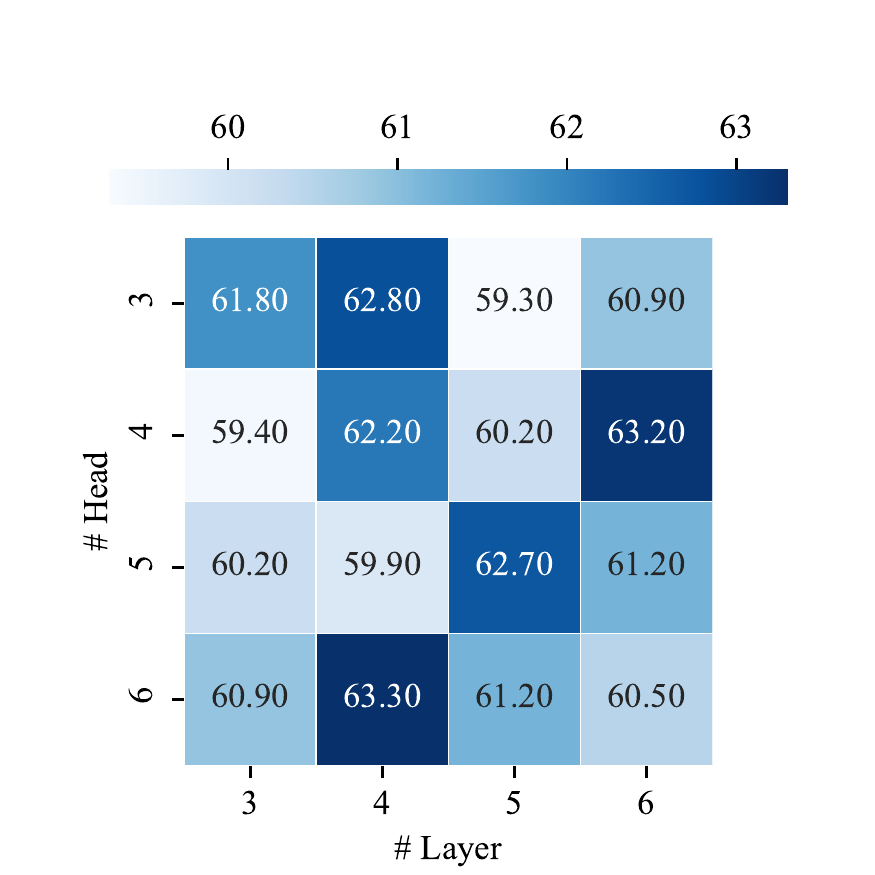}
		\subcaption{The performance on the test split.}
		\label{heatmap_test}
	\end{minipage}
	
	\caption{The performances on the different margin $\gamma$.}
	\label{heatmap}
\end{figure}

Further, we present the visualization of different selections for 
hyper-parameters. All the hyper-parameters are selected in the full-data 
setting, since the zero-shot splits are diverse and hard to unify. The first 
one is the number of layers and the number of heads in the graph transformer 
network, shown in Fig. \ref{heatmap}. To make a comprehensive comparison, we 
search both the head number and the layer number within \{3,4,5,6\}. As can be 
seen from the heatmap, the darker color represents the higher performance. TaCo 
reaches the optimal performance when the head number is 6 and the layer number 
is 4.

In addition, we present the model performances under different margins 
$\gamma$. The search scope of $\gamma$ is within \{8,10,12,14,16\}. And we 
select the optimal performance on the test split. The optimal value of $\gamma$ 
is 12.

\vspace{-0.3cm}
\begin{figure}[t]
	\centering
	
	\begin{minipage}{\linewidth}
		\centering
		\includegraphics[scale=0.5]{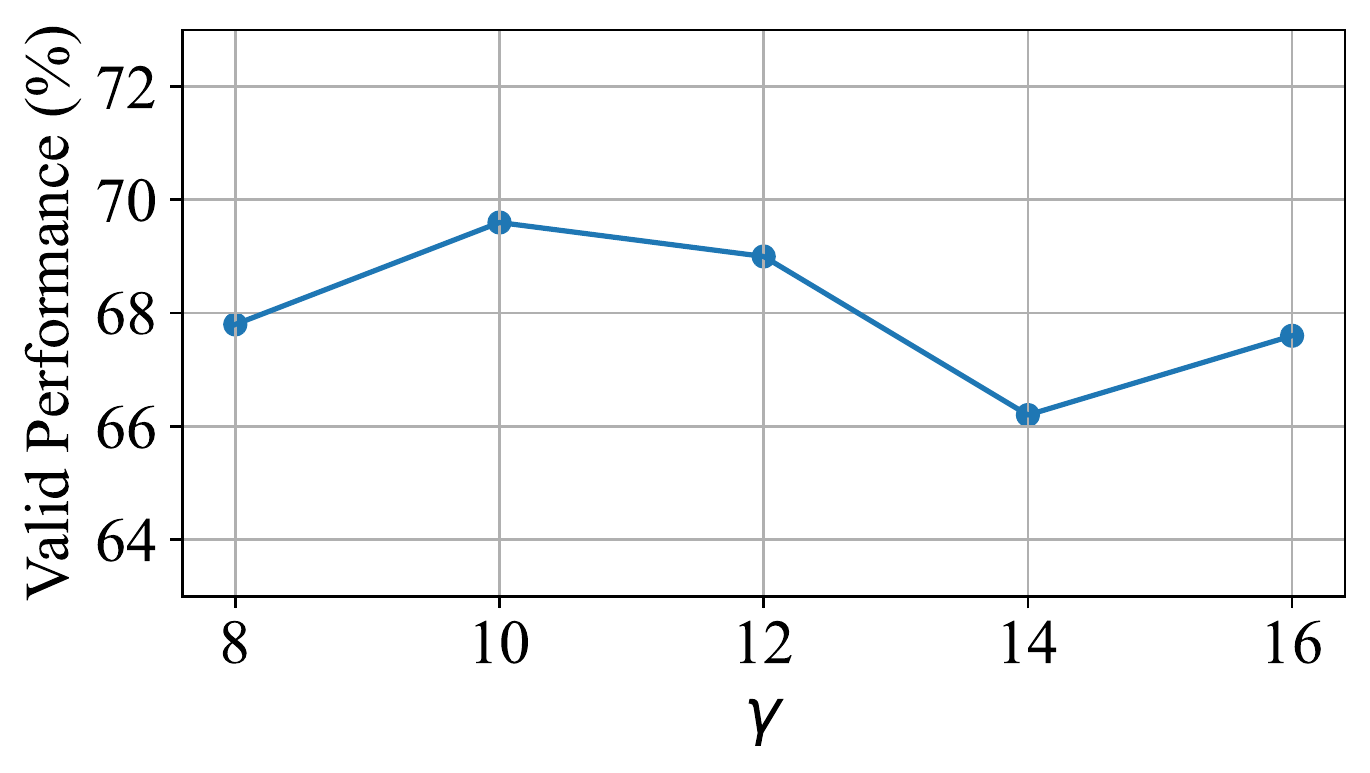}
		\subcaption{The performance on the validation split.}
		\label{gamma_valid}
	\end{minipage}
	
	\begin{minipage}{\linewidth}
		\centering
		\includegraphics[scale=0.5]{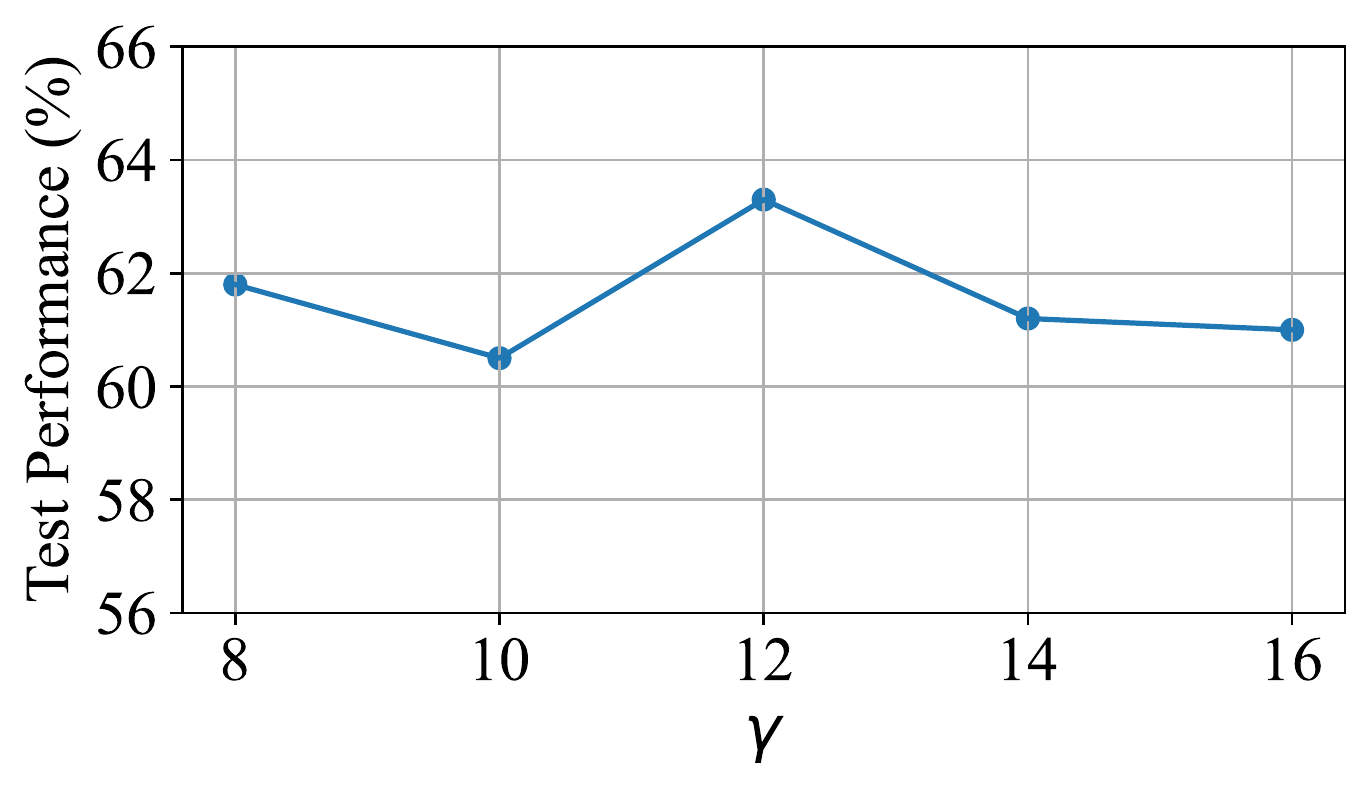}
		\subcaption{The performance on the test split.}
		\label{gamma_test}
	\end{minipage}
	
	\caption{The performances on the different margin $\gamma$.}
	\label{gamma}
\end{figure}

\section{Analysis of Model Generalization}
In this section, we will discuss the generalization capability of the proposed 
model. Since TaCo has achieved superior performances on the zero-shot setting, 
we further extend it to the full-data setting. Experiments are conducted on two 
mainstream logical reasoning dataset, ReClor and LogiQA. In Fig. 
\ref{tab:RECLOR_full}, we present the comparison results.

\subsection{Generalization on Full-data Setting of ReClor}

\begin{table}[t]
	\centering
	\caption{Experimental results on ReClor with the full-data 
		setting. \emph{Test-E} and \emph{Test-H} denote the easy part and the 
		hard part of the ReClor test split respectively.}
	\begin{tabular}{|p{3.5cm}|cccc|}
		\hline
		\multirow{2}{*}{\textbf{Model}} & 
		\multicolumn{4}{c|}{\textbf{ReClor}}\\ 
		&Dev &Test &Test-E &Test-H\\
		\hline
		\hline
		Random &25.00 &25.00 &25.00 &25.00\\
		Human Performance \cite{yu2019reclor} &- &63.00 &57.10 &67.20\\
		BERT-Large \cite{yu2019reclor} &53.80&49.80&72.00&32.30\\
		XLNet-Large \cite{yu2019reclor} &62.00 &56.00 &75.70 &40.50 \\
		RoBERTa-Large \cite{yu2019reclor}&62.60&55.60&75.50&40.00 \\
		DAGN \cite{huang2021dagn} &65.80 & 58.30 &75.91 &44.46  \\
		FocalReasoner \cite{ouyang2021fact} &66.80 &58.90 &77.05 &44.64\\
		LReasoner \cite{wang2021logic} & 66.20 &62.40 &81.40 
		&47.50 \\
		MERIt \cite{jiao2022merit}& 66.80 & 59.60 & 78.10 & 45.20 \\
		AdaLoGN \cite{li2022adalogn} &65.20 &60.20 &79.32 &45.18 \\
		Logiformer \cite{xu2022logiformer} & \underline{68.40} & 
		\textbf{63.50} &\underline{79.09} &\textbf{51.25} \\
		\hline
		\hline
		TaCo (Ours) & \textbf{69.00} & \underline{63.30} &\textbf{81.36} 
		&\underline{49.11} \\
		\hline
	\end{tabular}
	
	\label{tab:RECLOR_full}
\end{table}

Compared with all the current methods, TaCo achieves competitive results. In 
comparison with the SOTA model Logiformer, TaCo is better on the metric of 
\emph{Dev} and \emph{Test-E} with 0.60\% and 2.27\% gains. And it is only 
slightly lower than SOTA with 0.20\% in the test split. We argue that in the 
full-data setting, the ideal distribution of types leads to the weakening of 
the type modeling. Combined with previous experiments on the zero-shot setting, 
though Logiformer does good at over-fitting the full-data setting, it loses 
some generalization capability. From this perspective, TaCo also shows great 
superiority on the model generalization, performing consistently for both 
settings.

\begin{table*}[t]
	\centering
	\caption{Ablation Studies on the full-data setting. The drops on the 
		performances are marked in red. The darker ones represent the 
		greater decline in values.}
	
	\begin{tabular}{|p{6.2cm}|cccccc|}
		\hline
		\multirow{2}*{\textbf{Model}} &\multicolumn{6}{c|}{\textbf{ReClor}} \\
		~ &\textbf{Valid} &$\Delta$ &\textbf{Test} &$\Delta$ &\textbf{Test-E} 
		&\textbf{Test-H}\\
		\hline
		\hline
		TaCo (Ours) &69.00 & - &63.30 & - &81.36 &49.11 \\
		\textbf{a) Heuristic Input Reconstruction} & & & & & & \\
		\quad\quad w/o type prefix &67.60 &\cellcolor{red!15}-1.40 &61.30 
		&\cellcolor{red!25}-2.00 
		&80.45 &46.25 \\
		\quad\quad w/o input reconstruction  &67.80 &\cellcolor{red!15}-1.20 
		&61.10 &\cellcolor{red!25}-2.20 &79.09 &51.25  \\
		\textbf{b) Graph Construction \& Reasoning} & & & & & & \\
		\quad\quad w/o graph reasoning &64.40 &\cellcolor{red!45}-4.60 &59.30 
		&\cellcolor{red!45}-4.00 &79.55 &43.39  \\
		\quad\quad w/o global node &65.60 &\cellcolor{red!35}-3.40 &61.90 
		&\cellcolor{red!15}-1.40 &81.82 &46.25  \\
		\textbf{c) Type-aware Contrastive Learning} & & & & & & \\
		\quad\quad w/o type contrast &67.80 &\cellcolor{red!15}-1.20 &62.90 
		&\cellcolor{red!10}-0.40 &79.55 &49.82 \\
		\hline
	\end{tabular}
	
	\label{ablation_full}
\end{table*}

Further we conduct the ablation studies to analyze the effectiveness of the 
proposed modules in the full-data setting. From the results, the graph 
reasoning contributes most to the performance on the test split. Also, we only 
witness 0.40\% improvements for the consideration of type-aware margin loss. 
Since in the full-data setting, the train and test splits share the same 
distributions, the effects of the reasoning types are reduced. Therefore, the 
slight gain of the contrastive learning is reasonable. In all, the 
generalization capability of TaCo is well verified.

\subsection{Generalization on Other Dataset}
To discuss the generalization capability of TaCo, we further conduct some 
experiments on another logical reasoning dataset LogiQA\cite{liu2021logiqa}. 
Following the method utilized for ReClor, we also label each example of LogiQA 
with one of the 17 reasoning types, and consider the zero-shot setting. We 
apply the above mentioned strategies to obtain three zero-shot splits (i.e., 
z1, z2 and z3) for LogiQA.

In comparison, we take some previous logical reasoning models into account, 
including two typical language model BERT-Large and RoBERTa-Large, the 
graph-based method DAGN and previous SOTA model Logiformer. We rerun the 
zero-shot setting for each baseline.

\begin{table}[t]
	\centering
	\caption{Experiments results on the zero-shot splits of the LogiQA dataset.}
	\scriptsize
	\begin{tabular}{|p{1.4cm}|cc|cc|cc|}
		\hline
		\multirow{2}*{\textbf{Model}} &\multicolumn{2}{c|}{\textbf{z1}} 
		&\multicolumn{2}{c|}{\textbf{z2}} &\multicolumn{2}{c|}{\textbf{z3}} \\
		~ &\textbf{Test-A} &\textbf{Test-U} &\textbf{Test-A} 
		&\textbf{Test-U}&\textbf{Test-A}&\textbf{Test-U}\\
		\hline
		\hline
		BERT-L &26.26 &28.64 &27.80 &27.06 &28.26 &27.14 \\
		RoBERTa-L &27.65 &33.98 &29.49 &31.76 &28.73 &29.43 \\
		DAGN &34.41 &\textbf{40.78} &\underline{35.48} &\underline{34.12} 
		&33.79 &32.00\\
		Logiformer &\underline{35.33} &36.41 &33.18 &\textbf{35.29} 
		&\underline{34.56} 
		&\underline{33.43}\\
		\hline
		\hline
		TaCo (Ours) &\textbf{35.58} &\underline{37.38} &\textbf{36.76} 
		&\textbf{35.29} &\textbf{36.41} &\textbf{35.14} \\
		\hline
	\end{tabular}
	
	\label{other_dataset}
\end{table}

From the results shown in TABLE \ref{other_dataset}, TaCo shows its superiority 
in most cases, only except the \emph{Test-Unseen} metric of z1 split. 
Specifically, compared with previous SOTA Logiformer, TaCo achieves an average 
gain of 1.89\% and 0.89\% on the \emph{Test-All} and \emph{Test-Unseen} 
respectively. In general, experiments on the LogiQA dataset prove the 
good generalization capability of TaCo, which satisfies our expectations.

\section{Case Study}
\begin{figure}[t]
	\centering
	\begin{minipage}{\linewidth}
		\centering
		\includegraphics[scale=0.9]{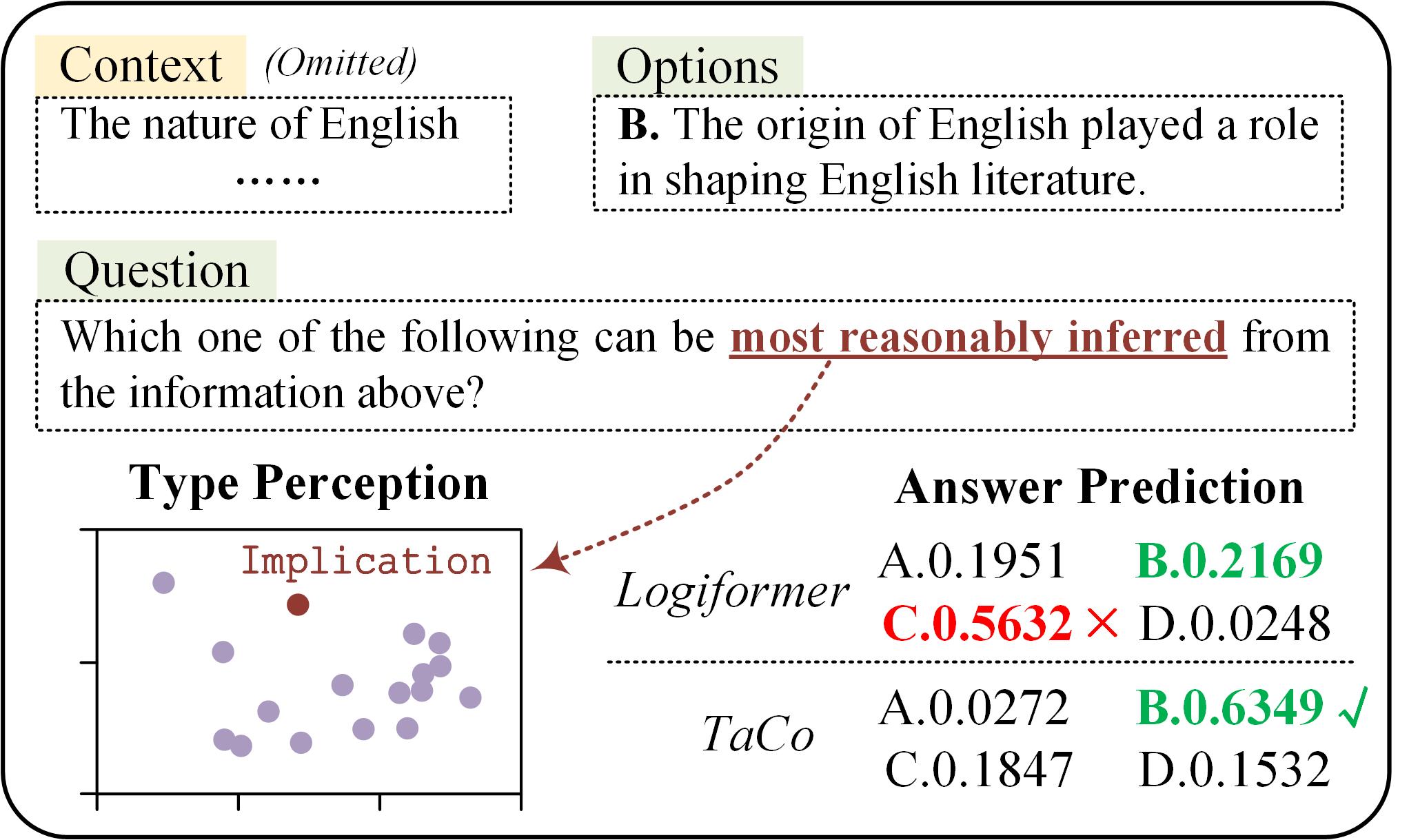}
		\subcaption{A successful case.}
		\label{success}
	\end{minipage}
	
	\begin{minipage}{\linewidth}
		\centering
		\includegraphics[scale=0.9]{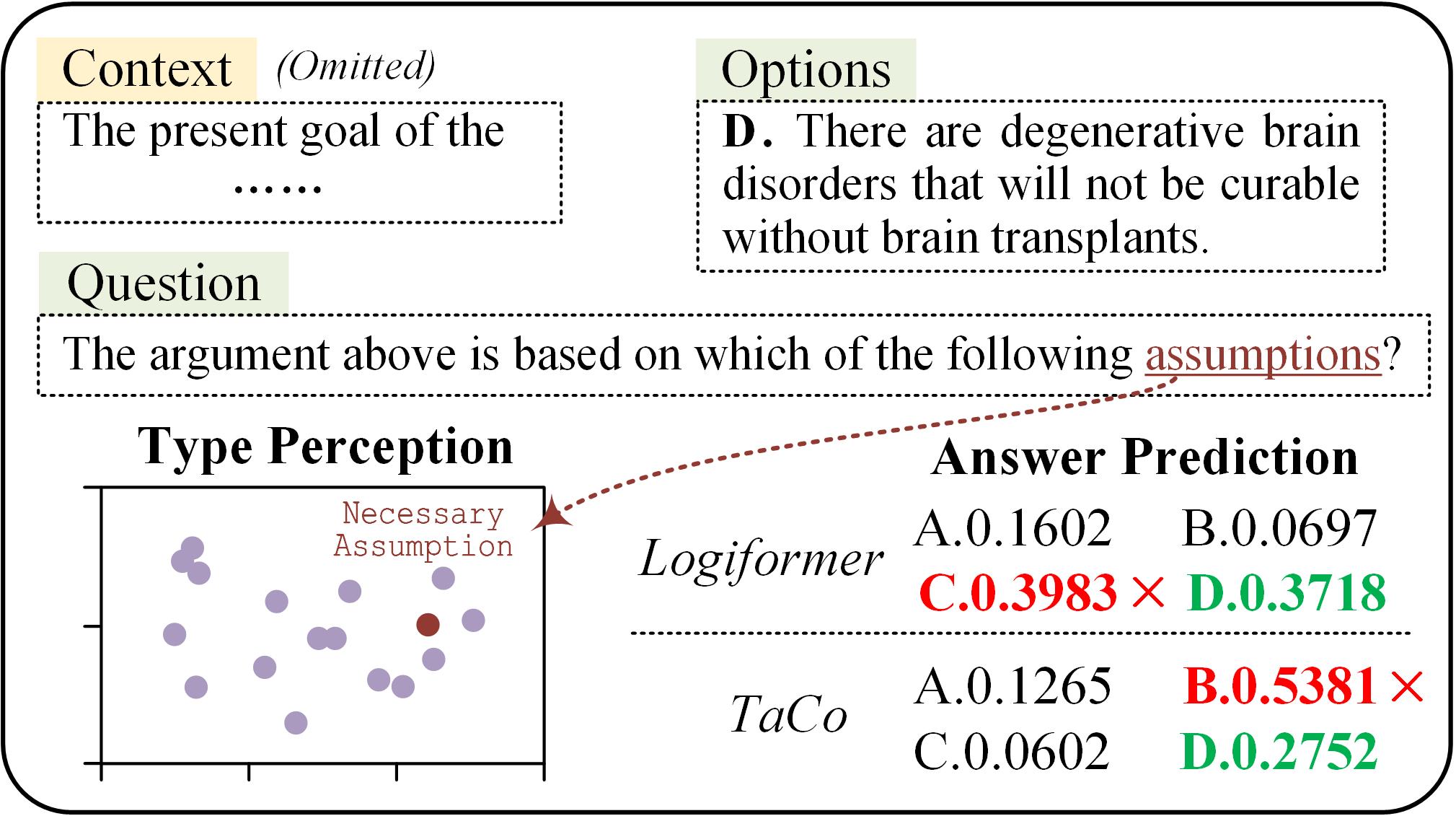}
		\subcaption{A failure case.}
		\label{failure}
	\end{minipage}
	\caption{Case Study.}
	\label{case_study}
\end{figure}

In this section, we discuss the interpretability of TaCo for the perception of 
reasoning types. Fig. \ref{case_study} shows a successful case and a 
failure case. For each case study, we present the visualization of the type 
perception via dimension reduction and make the comparison with the SOTA model 
Logiformer. For the successful case, TaCo can well distance the ground truth 
type \emph{Implication} with others and thus make the correct prediction. In 
this case, Logiformer which lacks the type perception fails. For the failure 
one, TaCo obviously struggles at distinguishing \emph{Necessary assumption} 
from other reasoning types. It leads to the wrong prediction, which is in 
same with Logiformer. It encourages us to explore the deeper modeling of 
types from the graph construction. In all, the type-aware contrastive 
learning is helpful to the answer prediction and provides the 
interpretability of the model.

\section{Conclusion and Future Work}

To study the zero-shot capability of the logical reasoning models, we propose 
the first benchmark for the generalized zero-shot logical reasoning, named 
ZsLR. It includes six splits sampled with three strategies and two 
metrics to comprehensively evaluate the performances. Also, we propose 
a model TaCo to enhance the reasoning type perception through the heuristic 
input reconstruction and the type-aware contrastive learning. Also, we conduct 
extensive experiments on the zero-shot splits, full-data setting as well as other dataset. Superior results illustrate the effectiveness and generalization capability of the proposed modules.

In the future, we encourage more works to mind reasoning manners in the logical 
reasoning task. And we are also interested in exploring the data-augmentation methods through logics.


\bibliographystyle{IEEEtran}
\bibliography{custom}

\end{document}